\def\year{2021}\relax
\documentclass[letterpaper]{article} %
\usepackage{aaai21}  %
\usepackage{times}  %
\usepackage{helvet} %
\usepackage{courier}  %
\usepackage[hyphens]{url}  %
\usepackage{graphicx} %
\usepackage{graphicx}  %
\usepackage{natbib}  %
\usepackage{caption} %
\usepackage{algorithm}
\usepackage{algpseudocode}
\usepackage{amsmath}
\usepackage{amssymb}
\usepackage{gensymb}
\usepackage{booktabs}
\usepackage{multirow}
\usepackage{subcaption}
\usepackage{diagbox}
\DeclareMathOperator*{\argmin}{argmin}
\usepackage[switch]{lineno}
\usepackage{fancyhdr}
\title{Generating Adversarial yet Inconspicuous Patches with a Single Image}
\author{

    Jinqi Luo, Tao Bai\thanks{Corresponding author.}, Jun Zhao

}
\affiliations{
    Nanyang Technological University, 50 Nanyang Avenue, Singapore, 639798\\
    luoj0021@ntu.edu.sg, bait0002@ntu.edu.sg, junzhao@ntu.edu.sg

}

\begin{document}

\maketitle
\thispagestyle{fancy}
\pagestyle{fancy}
\lhead{In Proceedings of the AAAI Conference on Artificial Intelligence 2021 Student Abstract and Poster Program}
\rhead{}
\cfoot{}
\renewcommand{\headrulewidth}{0.4pt}
\renewcommand{\footrulewidth}{0pt}

\begin{abstract}
\begin{quote}
Deep neural networks have been shown vulnerable to adversarial patches, where exotic patterns can result in model’s wrong prediction. Nevertheless, existing approaches to adversarial patch generation hardly consider the contextual consistency between patches and the image background, causing such patches to be easily detected by human observation. Additionally, these methods require a large amount of data for training, which is computationally expensive. To overcome these challenges, we propose an approach to generate adversarial yet inconspicuous patches with one single image. In our approach, adversarial patches are produced in a coarse-to-fine way with multiple scales of generators and discriminators. The selection of patch location is based on the perceptual sensitivity of victim models. Contextual information is encoded during the Min-Max training to make patches consistent with surroundings. 

\end{quote}

\end{abstract}

\section{Introduction}
In recent years, adversarial patch-based attack \cite{Brown2017AdversarialP} are proposed. %
However, existing adversarial patches are usually ended being noticeable for the human observer because of their exotic appearance. 
In addition, existing methods~\cite{Brown2017AdversarialP, Liu2019PerceptualSensitiveGF} require a large amount of quality data for training, which is computationally expensive and time-consuming.
Towards bridging research gaps mentioned above, we propose a GAN-based approach to generate Adversarial yet Inconspicuous Patches (AIP) trained from one single image.
Our approach captures the most sensitive area of the victim image, and applies adversarial patches generated with well-crafted objective functions. 
The goals of AIP are (1) pioneering in crafting adversarial patches with only one image, and (2) evading human detection while keeping attacks successful.

\begin{figure}[!h]

    \centering
    \includegraphics[width=0.95\columnwidth]{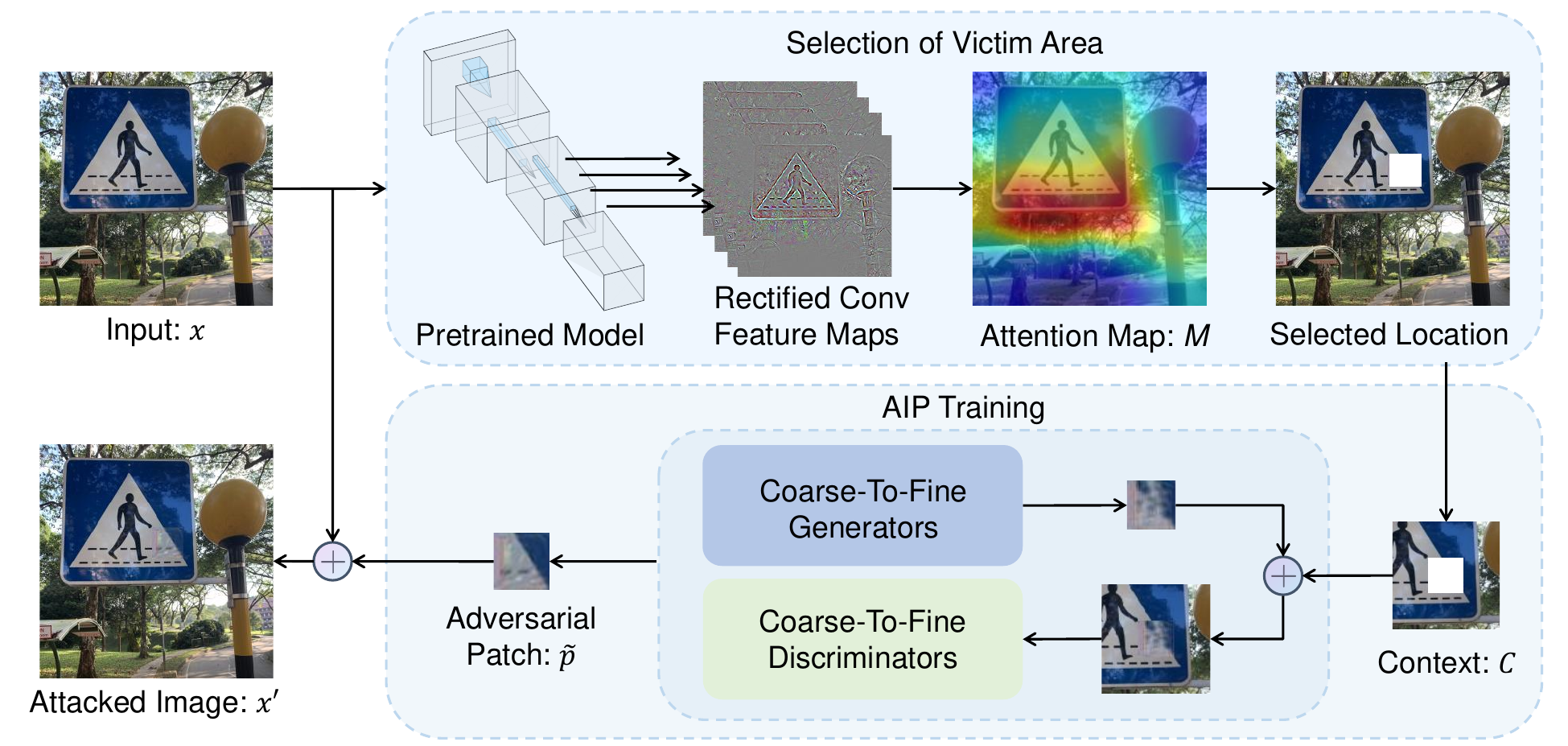}
    \caption{The overall framework of our approach.}
    \label{approach}
    \vspace{-0.15in}
\end{figure}

\section{Adversarial yet Inconspicuous Patches}

\subsection{AIP Framework}
\begin{figure}[h]

    \centering
    \includegraphics[width=0.85\columnwidth]{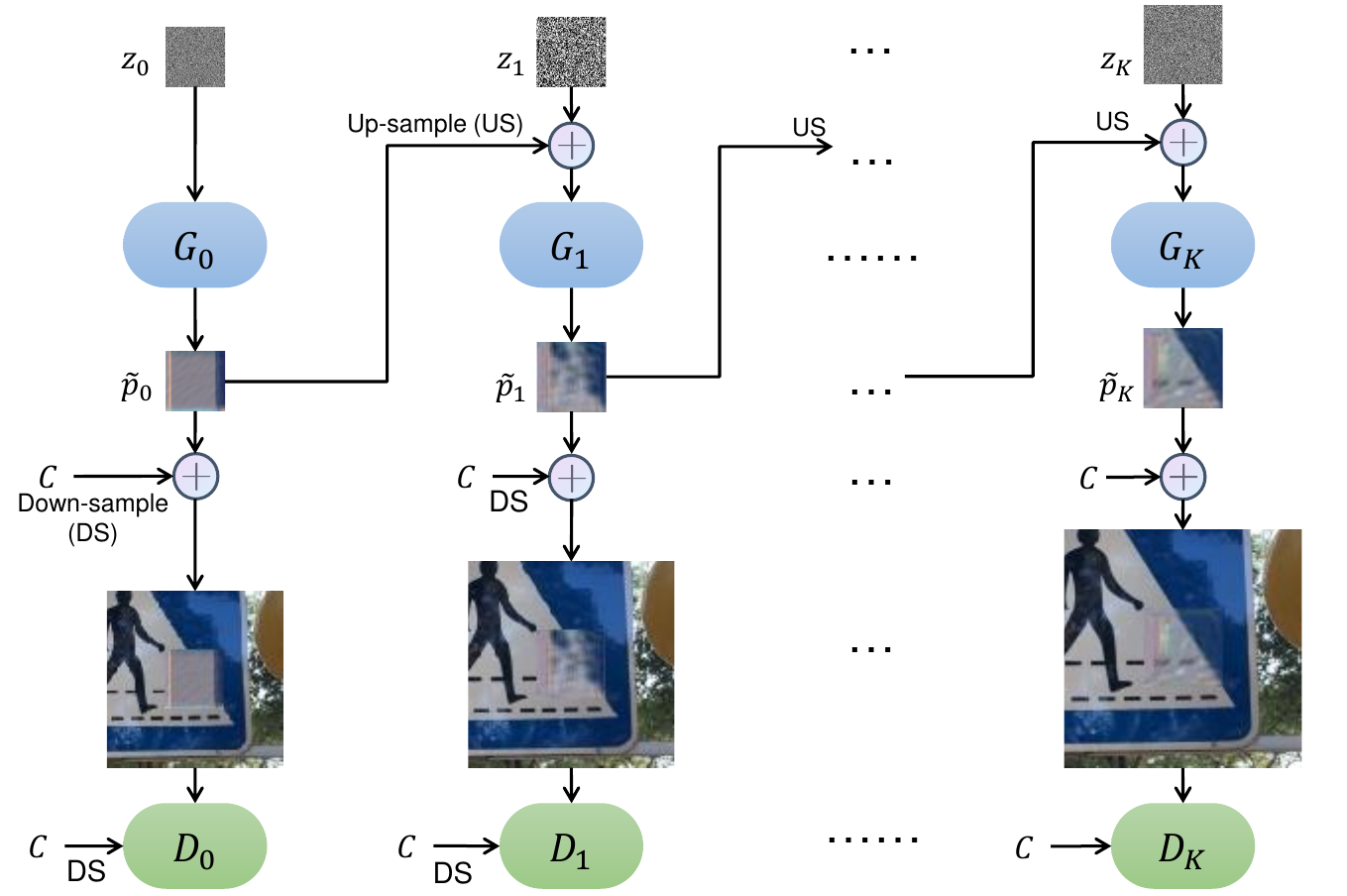}
    \caption{Structure of coarse-to-fine pipeline.}
    \label{AIP_structure}
    \vspace{-0.15in}
\end{figure}

\begin{figure}[h]
    \centering
    \includegraphics[width=\columnwidth]{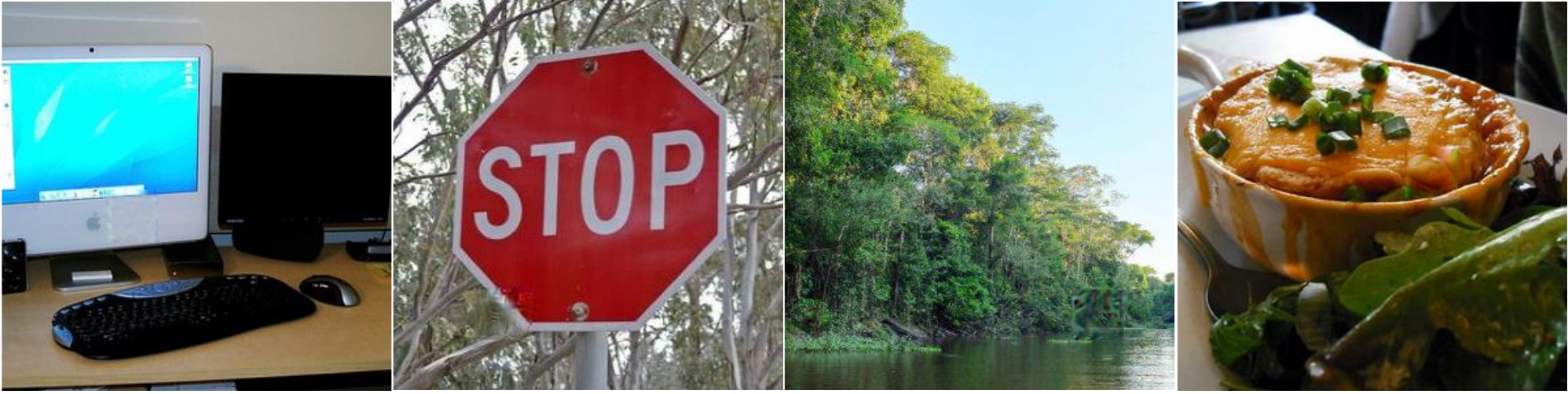}
    \caption{Some AIP examples. At first glance, most of the our adversarial patches are inconspicuous to observers.}
    \label{sample demo}
\end{figure}

The overview of our framework is illustrated in Figure~\ref{approach}. Given a target image, an attention map with target model is generated to capture the model's sensitivity and decide the patch position.
Then we deploy a series of generator-discriminator pairs $\{(G_0, D_0),\dots,(G_K,D_K)\}$, where $K$ is the total number of scales in the structure shown in Figure~\ref{AIP_structure}. These generator-discriminator pairs are trained against an image pyramid of $p$ and $C$. 
Correspondingly, the image pyramid is expressed as $\left\{(p_0,C_0)\dots(p_K,C_K)\right\}$, where $p_i$ and $C_i$ are downsampled version of $p$ and $C$ with a factor $r^{K-i}$ $(0<r<1)$.
In every scale, we execute adversarial training for generators and discriminators.
The generator $G_i$ is expected to produce realistic patches, and the discriminator attempts to distinguish generated samples from $p_i$.
Since our approach requires the generated patches to be consistent with original images,
the input of discriminator is the surrounding context $C_i$ with the intermediate patches $p_i$ placed right at the center of context.
During training, the background information will be encoded to the generator progressively. Some examples of generated AIP are shown in Figure~\ref{sample demo}.

\subsection{Objective Functions}
We take the $i_{th}$ scale to elaborate the training details. We denote the output of $G_{i-1}$ as $\tilde{p}_{i-1}$, then the input for $G_i$ is 
\begin{equation}\tilde{p}_{i}=G_{i}\left(z_{i},\left(\bar{p}_{i-1}\right) \uparrow^{r}\right), \end{equation}
where $\bar{p}_{i-1} \uparrow^{r}$ is the upsampled patch of $\tilde{p}_{i-1}$.

The GAN adversarial loss %
can be written as
\begin{equation}
\begin{aligned}
\mathcal{L}_{\mathrm{GAN}} = &\mathbb{E}_{p_i \sim x} \log \mathcal{D}(p_{i}, C_i)+ \\ 
 & \mathbb{E}_{z_i \sim \mathcal{P}_z} \log (1-\mathcal{D}(\mathcal{G}(z_i, \tilde{p}_{i-1}), C_i)),
\end{aligned}
\end{equation}
where $\mathcal{P}_z$ is a prior for noises.
The loss for fooling target model $f$ in untargeted attacks is
\begin{equation}
\mathcal{L}_{\mathrm{adv}}^{f}=\mathbb{E}_{x} \ell_{f}(x \oplus \bar{p}_{i} \uparrow^{r}, y),
\end{equation}
where $\ell_{f}$ denotes the loss function used in the training of $f$, and $y$ is the true class of $x$.

To stabilize the training of GAN, we add the reconstruction loss
\begin{equation}
\mathcal{L}_{\mathrm{rec}}=\left\|G_{i}\left(z_i, \tilde{p}_{i-1}\right)-p_{i}\right\|^{2}.
\end{equation}

We also add a total variation loss
 \begin{equation}
 \begin{aligned}
 \mathcal{L}_{\mathrm{tv}} = \sum_{a=0}^h \sum_{b=0}^w (\left|p_i^{(a+1,b)}-p_i^{(a,b)}\right|+\left|p_i^{(a,b+1)}-p_i^{(a,b)}\right|)
 \end{aligned}
 \end{equation}
as a regularization term to ensure that the texture of generated patches is smooth enough.
Finally, the full objective function in $i_{th}$ scale can be expressed as
\begin{equation}\mathcal{L}=\mathcal{L}_{\text {adv }}^{f}+\alpha \mathcal{L}_{\mathrm{GAN}}+\beta \mathcal{L}_{\mathrm{rec}}+\gamma \mathcal{L}_{\mathrm{tv}},
\label{eqn:overall loss}
\end{equation}
where $\alpha$, $\beta$ and $\gamma$ are to balance the relative importance of each loss.
Then we train our generator and discriminator by solving the min-max game as 
\begin{equation}
\argmin_{G_{i}} \max _{D_{i}} \mathcal{L}\left(G_{i}, D_{i}\right).
\end{equation}

\section{Experiment Results}

\subsection{White-box and Black-box Attack}
To assess the attack capability of the adversarial patches generated, we conduct experiments in white-box setting and black-box setting respectively.
Our data are randomly sampled from ImageNet.
Due to resource limitation, we first choose 10 classes from ImageNet, and sample 10 images in each class. For each image, 1000 patches will be generated.
Results are shown in Table~\ref{success rate}.

\begin{table}[h]
\centering
\fontsize{9}{9}\selectfont
\begin{tabular}{lccccc}
\toprule 

 & White-box & \multicolumn{4}{c}{Black-box} \\ \cmidrule(r){2-2} \cmidrule(r){3-6}
       & Inception & Google & MNAS & Mobile & L2-Mobile \\ \midrule
\textbf{Persian Cat}      & 100.00\%             & 99.22\%            & 85.62\%          & 90.66\%          & 80.33\%            \\ %
\textbf{Zebra}            & 98.53\%              & 99.36\%            & 85.58\%          & 90.38\%          & 74.40\%            \\ %
\textbf{Balloon}          & 99.52\%              & 99.19\%            & 79.68\%          & 90.19\%          & 82.70\%            \\ %
\textbf{Desktop} & 99.72\%              & 99.20\%            & 82.23\%          & 90.71\%          & 77.86\%            \\ %
\textbf{Table}            & 99.95\%              & 99.19\%            & 86.13\%          & 90.09\%          & 87.09\%            \\ %
\textbf{Hourglass}        & 99.99\%              & 99.20\%            & 82.12\%          & 90.59\%          & 80.39\%            \\ %
\textbf{Truck}     & 99.97\%              & 99.31\%            & 85.34\%          & 91.06\%          & 75.65\%            \\ %
\textbf{Street Sign}      & 98.30\%              & 99.21\%            & 82.27\%          & 90.10\%          & 84.92\%            \\ %
\textbf{Potpie}           & 99.88\%              & 99.30\%            & 83.81\%          & 90.85\%          & 80.01\%            \\ %
\textbf{Lakeside}         & 99.91\%              & 99.30\%            & 84.00\%          & 89.96\%          & 71.85\%            \\ \midrule
\textbf{Average}              & 99.58\%              & 99.25\%            & 83.68\%          & 90.46\%          & 79.52\%            \\ \bottomrule
\end{tabular}

\caption{White-box and Black-box attack success rates. The victim model under white-box is InceptionV3 and the victims under black-box are GoogleNet, MNASNet (multipier of 1.0), MobileNetV2, and MobileNetV2 with L2 robust training ($\epsilon = 3$).}
\label{success rate}
\vspace{-0.15in}
\end{table}

\subsection{Human Observer Evaluation}
We evaluate the risks of adversarial patches prone to human detection. We compete our synthetic patches with Google Patch \cite{Brown2017AdversarialP} and PS-GAN \cite{Liu2019PerceptualSensitiveGF} while including original images as the baseline. Note that in each background image, all the patches are attached in the same location for fairness.
In total we collected 102 answer sheets and the rates of images that are labeled as patch-detected are summarized in Table~\ref{human identify}.
\begin{table}[H]
\centering
\resizebox{0.85\columnwidth}{!}{%
\begin{tabular}{ccccc}
\toprule
Natural Image & Google Patch  & PS-GAN  & AIP \\ \midrule
12.15\%       & 93.63\%    & 89.90\% & 36.96\% \\ \bottomrule
\end{tabular}%
}
\caption{Average percentage of images that users label them as \textit{Synthetic Patch Detected}.}
\label{human identify}
\vspace{-0.08in}
\end{table}

\section{Conclusion}
In this work, we propose an approach of GAN-based adversarial networks trained with only one image to produce adversarial patches.
Our approach employs multiple scales of generators with discriminators to generate patches in a coarse-to-fine way.
To equip our approach with stronger attacking capability, we consider the perceptual sensitivity of victim model by developing model attention mechanism.
Through extensive experiments, our approach shows satisfying attack capabilities, black-box transferabilities, and good performance to evade detection in human evaluation.

\bibliography{abstract_references.bib}
\end{document}